\documentclass{article}

\usepackage{PRIMEarxiv}

\usepackage[utf8]{inputenc} % allow utf-8 input
\usepackage[T1]{fontenc}    % use 8-bit T1 fonts
\usepackage{hyperref}       % hyperlinks
\usepackage{url}            % simple URL typesetting
\usepackage{booktabs}       % professional-quality tables
\usepackage{amsfonts}       % blackboard math symbols
\usepackage{nicefrac}       % compact symbols for 1/2, etc.
\usepackage{microtype}      % microtypography
\usepackage{lipsum}
\usepackage{fancyhdr}       % header
\usepackage{multirow}
\usepackage[table]{xcolor}
\usepackage{graphicx}   
\usepackage{amsmath}
\usepackage{comment}
\usepackage{float}  % preamble'da zaten varsa gerek yok
\usepackage[normalem]{ulem}

\AtBeginDocument{\raggedbottom}
\graphicspath{{media/}}     % organize your images and other figures under media/ folder

\title{CLoE: Curriculum Learning on Endoscopic Images for Robust MES Classification %in Ulcerative Colitis
}

\author{
  Zeynep Ozdemir \\
  Computer Engineering Department \\
  Ankara University \\
  zynpozdemir@ankara.edu.tr \\
   \And
  Hacer Yalim Keles \\
  Computer Engineering Department \\
  Hacettepe University \\
  hacerkeles@cs.hacettepe.edu.tr \\
  \And
  Omer Ozgur Tanriover \\
  Computer Engineering Department \\
  Ankara University \\
  tanriover@ankara.edu.tr \\
}

\begin{document}
\maketitle

\begin{abstract}
\noindent
Estimating disease severity from endoscopic images is essential in assessing ulcerative colitis, where the Mayo Endoscopic Subscore (MES) is widely used to grade inflammation. However, MES classification remains challenging due to label noise from inter-observer variability and the ordinal nature of the score, which standard models often ignore. We propose CLoE, a curriculum learning framework that accounts for both label reliability and ordinal structure. Image quality, estimated via a lightweight model trained on Boston Bowel Preparation Scale (BBPS) labels, is used as a proxy for annotation confidence to order samples from easy (clean) to hard (noisy). This curriculum is further combined with ResizeMix augmentation to improve robustness. Experiments on the LIMUC and HyperKvasir datasets, using both CNNs and Transformers, show that CLoE consistently improves performance over strong supervised and self-supervised baselines. For instance, ConvNeXt-Tiny reaches 82.5\% accuracy and a QWK of 0.894 on LIMUC with low computational cost. These results highlight the potential of difficulty-aware training strategies for improving ordinal classification under label uncertainty. Code will be released at \url{https://github.com/zeynepozdemir/CLoE}.

\end{abstract}


\section{Introduction}
\label{sec:introduction}

Ulcerative colitis (UC) is a chronic inflammatory bowel disease characterized by persistent inflammation of the colonic mucosa. Achieving and maintaining clinical recovery is the primary goal in UC management. Accurate and consistent assessment of disease activity is therefore critical for effective treatment planning \cite{feuerstein2014ulcerative}. In clinical practice, disease activity is commonly evaluated using the Mayo Endoscopic Score (MES), which grades mucosal inflammation from 0 to 3. However, the ordinal nature of MES leads to ambiguity between adjacent classes. Furthermore, the scoring process relies heavily on expert judgment, resulting in considerable inter- and intra-observer variability \cite{0_dataset_gorkem_polat_2022_5827695, 3_polat2023improving}. High-grade inflammation is often underrepresented in datasets, and moderate cases exhibit strong disagreement among experts. In addition, visual artifacts such as stool, blood, or technical limitations can degrade image quality and introduce noisy labels. As a result, MES classification is a challenging task that involves ordinal structure, class imbalance, and label noise sensitivity \cite{9_polat2025class, 11_lee2025artificial}.

To address these challenges, several approaches have been proposed. For example, Polat et al. \cite{1_polat2022class} designed a loss function to mitigate class imbalance, while Kadota et al. \cite{7_kadota2024deep} introduced a ranking-based approach using relative annotations to reduce label uncertainty. Lee et al. \cite{13_lee2025multi} modeled inter-class contextual relationships via multi-task learning, and others such as Xu \cite{5_xu2024ssl} and Margapuri \cite{10_margapuri2025diagnosis} adopted self-supervised learning to reduce the need for labeled data. However, most of these methods do not explicitly model the effects of image quality, label noise, or sample difficulty.

Curriculum learning differs from standard training by introducing data in stages rather than all at once. It typically begins with simpler cases and gradually progresses to more complex ones, guiding the learning process \cite{jimenez2022curriculum}. While not yet applied to MES directly, curriculum learning has shown effectiveness in related medical imaging problems. Jiménez-Sánchez et al. \cite{jimenez2022curriculum} developed a difficulty-aware curriculum for femur fracture classification; Ma et al. \cite{ma2024adaptive} proposed an active curriculum based on class frequency and reliability for chest X-rays; Abbas et al. \cite{abbas2025clog} employed intra-class granularity in visual patterns; and Burduja et al. \cite{burduja2021unsupervised} selected samples based on deformation degrees in image registration. These studies highlight curriculum learning’s potential in scenarios where visual complexity and label quality are key factors.

Motivated by these insights, we propose CLoE (Curriculum Learning on Endoscopy), a curriculum-based training framework for MES classification that incorporates sample difficulty derived from image quality. We first train a lightweight image quality classifier on Boston Bowel Preparation Scale (BBPS) labels from the HyperKvasir dataset \cite{borgli2020hyperkvasir}. While MES assesses inflammation severity, BBPS evaluates bowel cleanliness. Despite their clinical differences, BBPS offers a meaningful proxy for visual reliability. This enables estimation of difficulty scores without expert-defined uncertainty labels. 

%We apply the trained quality classifier to both LIMUC \cite{0_dataset_gorkem_polat_2022_5827695} and HyperKvasir \cite{borgli2020hyperkvasir} datasets, assigning difficulty levels to each MES-labeled sample. These scores guide a curriculum schedule in which samples are presented progressively from easy (clean) to hard (noisy).  Early training begins with high-quality, low-uncertainty samples and then includes harder, potentially noisy examples, which aligns with the progressive sampling paradigm in curriculum learning \cite{bengio2009curriculum,hacohen2019power}.
We apply the trained quality classifier to both LIMUC \cite{0_dataset_gorkem_polat_2022_5827695} and HyperKvasir \cite{borgli2020hyperkvasir}, assigning each MES-labeled sample a difficulty score. These scores define a curriculum schedule that initially prioritizes high-quality, low-uncertainty samples and gradually incorporates more challenging, noisier cases, following the progressive sampling strategy of curriculum learning \cite{bengio2009curriculum,hacohen2019power}.

%To further enhance generalization in low-resource (data scarce) and imbalanced scenarios, we integrate ResizeMix augmentation into our framework. This technique increases image variation while helping to preserve label and ordinal consistency. CLoE  is evaluated across multiple backbone families, including ResNet \cite{resnet_targ2016}, DenseNet \cite{densenet_huang2017densely},  MobileNetV2 \cite{sandler2018mobilenetv2}, ConvNeXt \cite{convnext_liu2022convnet}, EfficientNet \cite{efficientnet_tan2019}, and Vision Transformer (ViT) \cite{vit_dosovitskiy2020image} and generally improved performance in our experiments.
To further enhance generalization in low-resource and imbalanced scenarios, we integrate ResizeMix augmentation into our framework. This technique increases image variation while preserving label and ordinal consistency. We evaluate CLoE across diverse backbone families, including ResNet \cite{resnet_targ2016}, DenseNet \cite{densenet_huang2017densely}, MobileNetV2 \cite{sandler2018mobilenetv2}, ConvNeXt \cite{convnext_liu2022convnet}, EfficientNet \cite{efficientnet_tan2019}, and Vision Transformer (ViT) \cite{vit_dosovitskiy2020image}, and observe consistent performance gains across all experiments.

\vspace{0.5em}
\noindent The main contributions of this work are as follows:
\begin{itemize}
    \item We introduce \textbf{CLoE}, a curriculum learning framework that leverages image quality as a proxy for sample difficulty. To the best of our knowledge, this is the first work to apply curriculum learning to MES classification, enabling more reliable predictions without relying on expert-defined uncertainty scores.
    
    \item We design a three-phase, difficulty-aware training schedule that partitions pseudo-labeled samples into clean (high-confidence), mixed (moderate-confidence), and noisy (low-confidence) subsets. This structured progression is combined with ResizeMix augmentation to improve robustness against label noise while preserving ordinal consistency.
    
    \item We demonstrate the effectiveness of CLoE across diverse architectures, achieving state-of-the-art performance: 82.5\% accuracy and QWK = 0.894 on LIMUC, and 80.0\% accuracy and QWK = 0.739 on HyperKvasir. These results consistently outperform prior MES classification methods.
\end{itemize}

\begin{comment}
\begin{itemize}
    \item We introduce CLoE, a curriculum learning framework that leverages image quality as a proxy to model difficulty, enabling more reliable MES classification without using expert-defined uncertainty scores.
    
    \item We design a three-phase, difficulty-aware training schedule that groups pseudo-labeled samples into clean (high-confidence), mixed (moderate-confidence), and noisy (low-confidence) subsets based on estimated reliability. This structured progression is  combined with ResizeMix augmentation, which aims to increase robustness to label noise and to preserve ordinal consistency.
    
    \item CLoE is tested across a variety of architectures and achieves state-of-the-art results: 82.51\% accuracy and 89.35 QWK on LIMUC, and 80.02\% accuracy and 73.94 QWK on HyperKvasir. These results outperform prior MES classification methods by approximately +2.5\% accuracy and +1.5 QWK.
\end{itemize}

\end{comment}

\section{Related Work}
This section reviews prior work on MES classification and curriculum learning in medical image analysis. We discuss common challenges, existing solutions, and how our approach addresses current limitations.

\subsection{Deep Learning Based Approaches}%for Assessing Ulcerative Colitis Severity from Endoscopic Images}

Estimating disease severity in UC from endoscopic images has received growing attention as a clinical decision support task. Most deep learning-based methods focus on predicting the MES, a widely used ordinal scale for UC evaluation \cite{3_polat2023improving,11_lee2025artificial}. However, this task remains challenging due to the ordinal nature of MES, class imbalance across severity grades, limited labeled data, and label noise arising from image artifacts or expert disagreement. These issues can significantly affect model performance, especially when standard classification methods do not explicitly account for ordering or uncertainty.

To address these challenges, several methods have been proposed. Some studies use custom loss functions that incorporate semantic distances between MES classes, such as the class distance-weighted cross-entropy loss (CDW-CE)  \cite{1_polat2022class,9_polat2025class}. Others treat MES prediction as a regression problem to better capture inter-class transitions \cite{3_polat2023improving,4_li2023comparative}. Data-efficient learning strategies, including self-supervised learning (SSL) \cite{5_xu2024ssl,10_margapuri2025diagnosis} and federated learning \cite{6_al2024ulcerative}, have also shown promise in scenarios with limited annotations. In addition, relative ranking approaches have been used to reduce inter-annotator variability \cite{7_kadota2024deep,15_shiku2025ordinal}.

To handle class imbalance, some studies adopt multi-task learning frameworks that combine MES classification with auxiliary tasks, such as inflammation grading \cite{13_lee2025multi}. Others use multiple instance learning to make predictions based on sets of images from the same patient \cite{12_gutierrez2025ulcerative,14_zhang2025afr}. Attention mechanisms and saliency maps have also been incorporated to improve model interpretability in clinical settings \cite{11_lee2025artificial,17_sano2025explainable}.

%SSL methods in particular have become popular due to their ability to reduce annotation needs. Contrastive learning approaches such as SimCLR \cite{simclr_chen2020simple}, MoCoV2 \cite{mocov2_chen2003improved}, BYOL \cite{byol_grill2020bootstrap}, and SimSiam \cite{simsiam_chen2021exploring} have achieved strong results in general vision tasks. However, these methods can struggle with medical images due to limited intra-class variation. Semantically similar samples may be incorrectly treated as negatives, which can reduce representation quality. In addition, these methods do not explicitly model the ordinal structure or uncertainty caused by visual quality.

%For example, Xu et al. \cite{5_xu2024ssl} proposed a SimCLR-based pretraining strategy for MES classification. While the method reported high performance, it does not directly address label noise or ordinal relationships, which are particularly important for middle and severe MES classes.

%Given these limitations, there is a need for methods that not only improve representation learning but also consider sample difficulty and annotation reliability. In this study, we propose a curriculum learning framework that explicitly incorporates both the ordinal nature of MES and the visual quality of endoscopic images.

Self-supervised learning (SSL) methods have gained popularity for reducing annotation needs. Contrastive learning approaches such as SimCLR \cite{simclr_chen2020simple}, MoCoV2 \cite{mocov2_chen2003improved}, BYOL \cite{byol_grill2020bootstrap}, and SimSiam \cite{simsiam_chen2021exploring} have shown strong results in general vision tasks. However, transferring these methods to medical imaging is challenging. Limited intra-class variation can cause semantically similar samples to be treated as negatives, reducing representation quality. In addition, they do not explicitly account for the ordinal structure or the uncertainty introduced by variable image quality. In MES classification, Xu et al. \cite{5_xu2024ssl} demonstrated that SimCLR-based pretraining can improve performance, yet their approach still overlooks label noise and ordinal relationships, which are particularly important for distinguishing middle and severe MES grades.

Given these limitations, there remains a need for methods that not only improve representation learning but also incorporate sample difficulty and annotation reliability. In this study, we propose a curriculum learning framework that explicitly models both the ordinal nature of MES and the visual quality of endoscopic images.

\subsection{Curriculum Learning in Medical Image Analysis}

Curriculum learning (CL) is a training strategy that presents data to the model in an easy-to-hard order to improve convergence and generalization \cite{bengio2009curriculum}. Inspired by human learning processes, CL has been shown to be effective in scenarios involving noisy, limited, or complex data \cite{hacohen2019power}. In recent years, it has been increasingly applied in medical image analysis.

As in MES classification, medical imaging tasks often involve class imbalance, annotation noise, and varying image quality. CL can help address these challenges by controlling the order in which samples are introduced during training. Starting with clear and reliable samples, and gradually including more difficult or noisy ones, can support more stable learning and better decision boundaries \cite{hacohen2019power}.

Several CL-based approaches have been explored in medical imaging. Jiménez-Sánchez et al. \cite{jimenez2022curriculum} proposed a curriculum based on prediction uncertainty for femur fracture classification. Ma et al. \cite{ma2024adaptive} combined active learning and curriculum strategies to handle limited annotations in chest X-rays. Abbas et al. \cite{abbas2025clog} used intra-class visual granularity to design a curriculum, while Burduja \cite{burduja2021unsupervised} and Dun \cite{dun2024trustworthy} applied CL to image registration and domain adaptation tasks, respectively.

However, most existing studies focus on binary or discrete classification problems and do not explicitly model the ordinal structure or the effects of image quality. These factors are particularly relevant in MES classification, where both severity levels and sample reliability vary significantly.

To the best of our knowledge, no prior work has used a clinical image quality measure such as the BBPS to guide curriculum learning in MES classification. In this study, we adapt CL to this setting by introducing a quality-aware training schedule that considers both sample difficulty and label reliability, aiming to fill an important gap in the literature.

\begin{figure}[tb!]
    \centering
    \includegraphics[width=\linewidth]{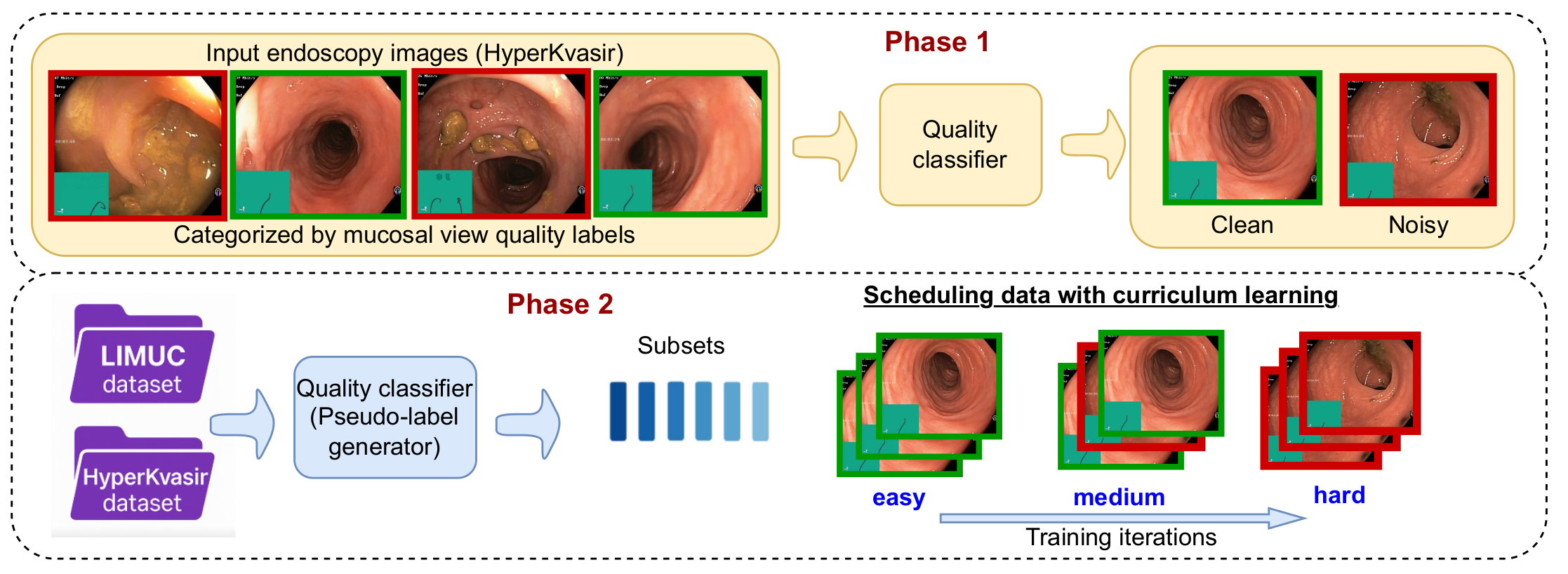}
    \caption{Overview of the proposed curriculum learning framework. A quality classifier is first trained on mucosal view labels from HyperKvasir. This model is then used to assign clean vs. noisy image scores to samples in both the LIMUC and HyperKvasir datasets. Based on these scores, training samples are divided into subsets of increasing difficulty and introduced to the MES classifier in a staged manner.}
    \label{fig:framework}
\end{figure}

\section{Methodology}
\label{sec:methodology}

%This section presents our proposed framework CLoE, developed for the task of MES classification. The framework comprises three key components operated in two phases: (1) estimating sample difficulty based on image quality, (2) designing a curriculum schedule to control the progressive introduction of training samples, and (3) implementing classifier training with multiple CNN and Transformer-based architectures. The process of our proposed cirriculum learning framework are depicted in Figure~\ref{fig:framework}.
This section introduces our proposed framework, CLoE, for MES classification. It consists of three main components applied in two phases: (1) estimating sample difficulty based on image quality, (2) designing a curriculum schedule to progressively introduce training samples, and (3) training classifiers with multiple CNN- and Transformer-based architectures. An overview of the proposed curriculum learning framework is shown in Figure~\ref{fig:framework}.

\subsection{Quality-Aware Difficulty Scoring}
\label{subsec:quality_aware}

A central component of curriculum learning is organizing training examples based on their difficulty. However, in medical imaging, difficulty is not directly measurable. To approximate it, we use image quality as a proxy, based on the assumption that clearer images are easier to interpret and more likely to have accurate labels. In contrast, low-quality images may obscure key visual patterns, increasing classification difficulty and annotation uncertainty.

This challenge is especially relevant for UC, where endoscopic images often suffer from visual artifacts due to bleeding, debris, or stool. These issues not only degrade visual clarity but can also introduce label noise, particularly in borderline cases. Presenting such ambiguous samples early during training may hinder generalization and lead to overfitting on noisy signals. A curriculum strategy mitigates these risks by first training on clean, reliable examples and gradually introducing harder cases. This progression promotes more stable and noise-resilient learning.

To implement this strategy, we train a binary image quality classifier using mucosal view annotations from the HyperKvasir dataset. Images with a BBPS score of 2–3 are labeled as \textit{clean}, while those with BBPS 0–1 or categorized as \textit{impacted stool} are labeled as \textit{noisy}. Figure~\ref{fig:limuc_samples} for visual samples.

We then use this classifier, which is based on a MobileNetV2 backbone, to assign a \textit{cleanliness probability} $s(x_i) \in [0,1]$ to each image in both the HyperKvasir and LIMUC datasets. These scores serve as sample-wise difficulty estimates and guide the curriculum schedule during MES classifier training, as detailed in Section~\ref{subsec:curriculum}.

%Further details on model training, dataset splits, and classifier evaluation are provided in Section~\ref{subsec:implementations}.

\subsection{Curriculum Strategy Design}
\label{subsec:curriculum}

Our goal is to organize the learning process in a way that reflects the gradual nature of human learning, leading to a more robust and balanced MES classifier. To this end, we design a simple but effective curriculum strategy that partitions the training data based on cleanliness scores obtained from the image quality classifier.

Each image \(x_i\) is assigned a cleanliness score \(s(x_i) \in [0,1]\), and the dataset \(D = \{(x_i, y_i)\}_{i=1}^{N}\) is divided into two subsets:

\[
D_{\text{clean}} = \{(x_i, y_i) \mid s(x_i) \ge \tau \}, \quad
D_{\text{noisy}} = \{(x_i, y_i) \mid s(x_i) < \tau \}
\]

We use a fixed threshold $\tau = 0.5$ to partition the training data into clean and noisy subsets, based on the cleanliness scores predicted by the image quality classifier. This binary split allows for structured curriculum stages by grouping high- and low-confidence samples. Our ablation results (see Section~\ref{subsec:ablation}) show that this staged training strategy leads to better model performance compared to training on the full dataset without separation.

Training is performed in three stages. First, the model is trained only on \(D_{\text{clean}}\) to learn from high-confidence, visually clear samples. Next, we use the combined set \(D_{\text{clean}} \cup D_{\text{noisy}}\).  Finally, the model is fine-tuned on \(D_{\text{noisy}}\) to adapt to more ambiguous samples.

We transition between these stages using early stopping. When validation accuracy does not improve for five epochs, training proceeds to the next phase. Details on sample proportions and stage transitions are provided in Section~\ref{sec:datasets_and_implementations}.

We also experimented with alternative curricula, including difficulty-weighted loss and sample reordering. However, because soft difficulty scores are often unreliable in noisy medical images, we found that binary subset–based scheduling produced more consistent results in our experiments.

\subsection{Evaluation Methodology and Architectural Details}%MES Classifier Architecture}
\label{subsec:mes_classifier}

To systematically evaluate the impact of different training paradigms on MES classification, we designed a three-stage comparison: (1) self-supervised learning (SSL), (2) conventional supervised learning, and (3) our proposed curriculum-based framework, CLoE. This design allows us to situate CLoE within a broader spectrum of strategies and quantify its relative contribution.

Unlike prior studies that focus primarily on novel SSL methods, we include representative SSL baselines to enable a fair cross-paradigm comparison. Specifically, we evaluate four well-known SSL approaches, namely SimCLR, SimSiam, BYOL, and MoCoV2, which are pre-trained on ImageNet with contrastive or predictive objectives and fine-tuned for MES classification using ResNet50 backbones.

For conventional supervised training and cirriculumbased learning, backbone selection was guided by a trade-off between model complexity, computational efficiency, and suitability for MES classification. To represent modern architectures, we included ConvNeXt-Tiny, a recent CNN design that incorporates Transformer-inspired components (e.g., large-kernel depthwise convolutions, LayerNorm, GELU), alongside ViT-Base as a Transformer-based counterpart. Classical supervised CNNs were represented by ResNet50 and DenseNet121, which remain widely adopted in medical imaging. We further incorporated SEResNet50, a ResNet variant with squeeze-and-excitation attention, to examine the impact of channel-wise attention mechanisms. For resource-constrained scenarios, we evaluated MobileNetV2 due to its lightweight design and efficient inference. EfficientNet-B4 was included with Noisy Student pretraining to assess the benefit of large-scale self-training for robustness. Finally, ViT-Base (MAE) was fine-tuned only in its final layers to retain general-purpose representations while adapting to MES classification. 
%We employ a diverse set of backbone architectures to ensure robustness across design choices: lightweight CNNs (MobileNetV2), standard residual networks (ResNet50, SEResNet50), dense convolutional networks (DenseNet121), efficient scaling models (EfficientNet-B4), and Transformer-inspired architectures (ConvNeXt-Tiny, ViT-Base).

To study the role of augmentation, we compare CutMix, MixUp, and ResizeMix. Among them, ResizeMix consistently improved performance, particularly in terms of QWK, and was therefore integrated into the curriculum-based schedule. Detailed augmentation analyses are provided in Section~\ref{subsubsec:augmentation_effect}.

\section{Datasets and Implementation Details}
\label{sec:datasets_and_implementations}

We evaluate the proposed CLoE framework using two endoscopic image datasets: LIMUC and HyperKvasir. LIMUC serves as the primary dataset for MES classification, while HyperKvasir is used both for training the image quality classifier and as an auxiliary benchmark. This section outlines dataset details and the experimental setup.

\subsection{LIMUC Dataset}
\label{subsec:limuc-dataset}

LIMUC comprises 19,537 endoscopic images collected from 1,043 colonoscopy procedures conducted at Marmara University between 2011 and 2019. Each image is annotated with a MES 0–3 by experienced gastroenterologists. The dataset includes an official image-level train-test split, from which we reserve 10\% of the training portion for validation. Note that splitting at the image level rather than the patient level may introduce potential data leakage. Our setup adheres to the protocol of Xu et al.~\cite{5_xu2024ssl}.

\begin{figure}[H]
\centering
\includegraphics[width=0.9\linewidth]{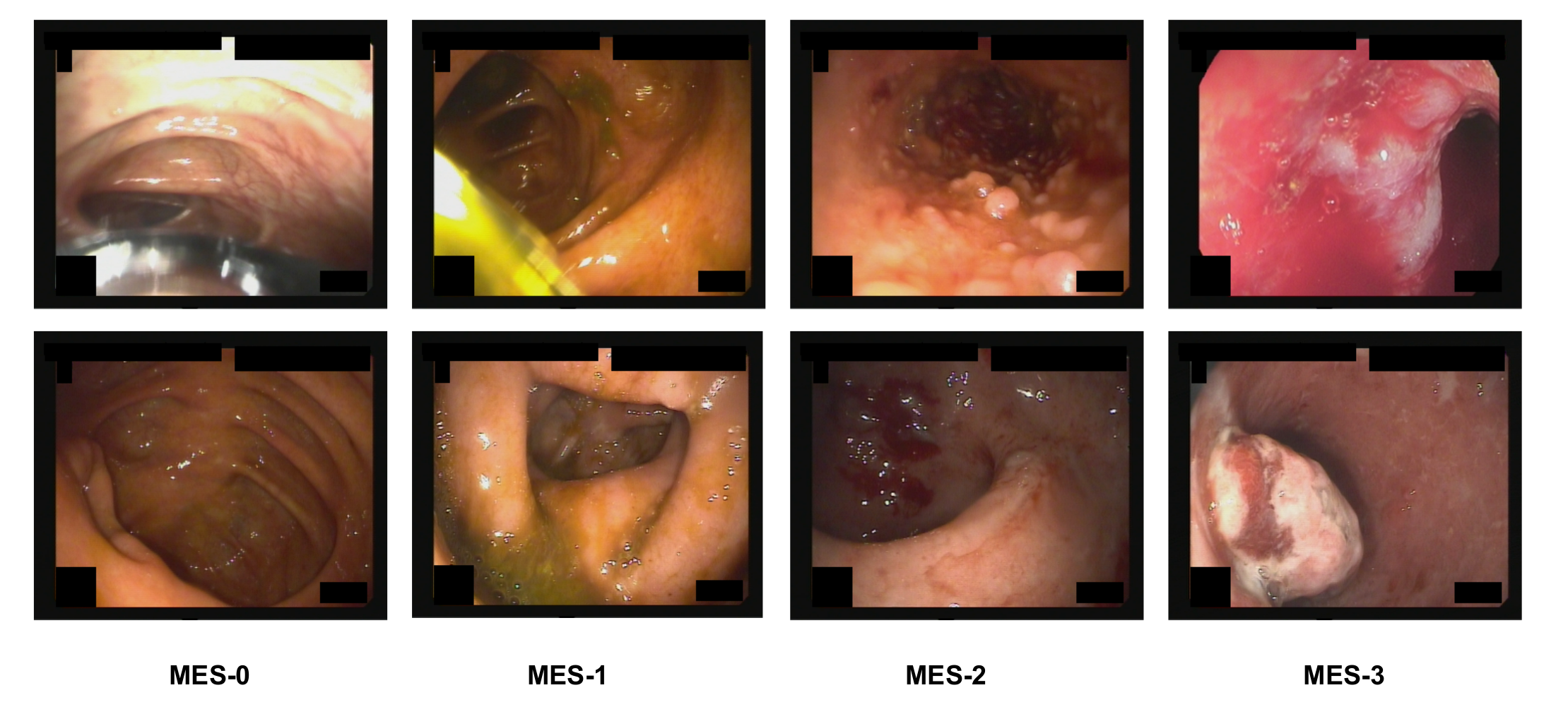}
\caption{Representative LIMUC images for each MES class.}
\label{fig:limuc_samples}
\end{figure}

Figure~\ref{fig:limuc_samples} presents representative LIMUC images for each MES grade, demonstrating the visual progression of inflammation severity.

\subsection{HyperKvasir Dataset}
\label{subsec:hyperkvasir-dataset}

HyperKvasir is a publicly available collection of gastrointestinal endoscopic images, used in our study for two main purposes: (1) training a binary image quality classifier and (2) training and evaluating MES classification models on an external dataset.

For the quality assessment task, we utilized a dedicated mucosal-view subset within HyperKvasir, annotated with BBPS scores and impacted stool labels. Images with BBPS scores of 2–3 were defined as clean (1,148 samples), and those with BBPS scores of 0–1 or labeled as impacted stool were considered noisy (776 samples).

For MES classification, we followed the selection protocol of Dermyer et al.~\cite{dermyer2025endodino}, using 777 images labeled with MES 1–3. The complete MES class distributions for LIMUC and HyperKvasir are summarized in Table~\ref{tab:mes_class_distribution_combined}.

\begin{figure}[H]
\centering
\includegraphics[width=\linewidth]{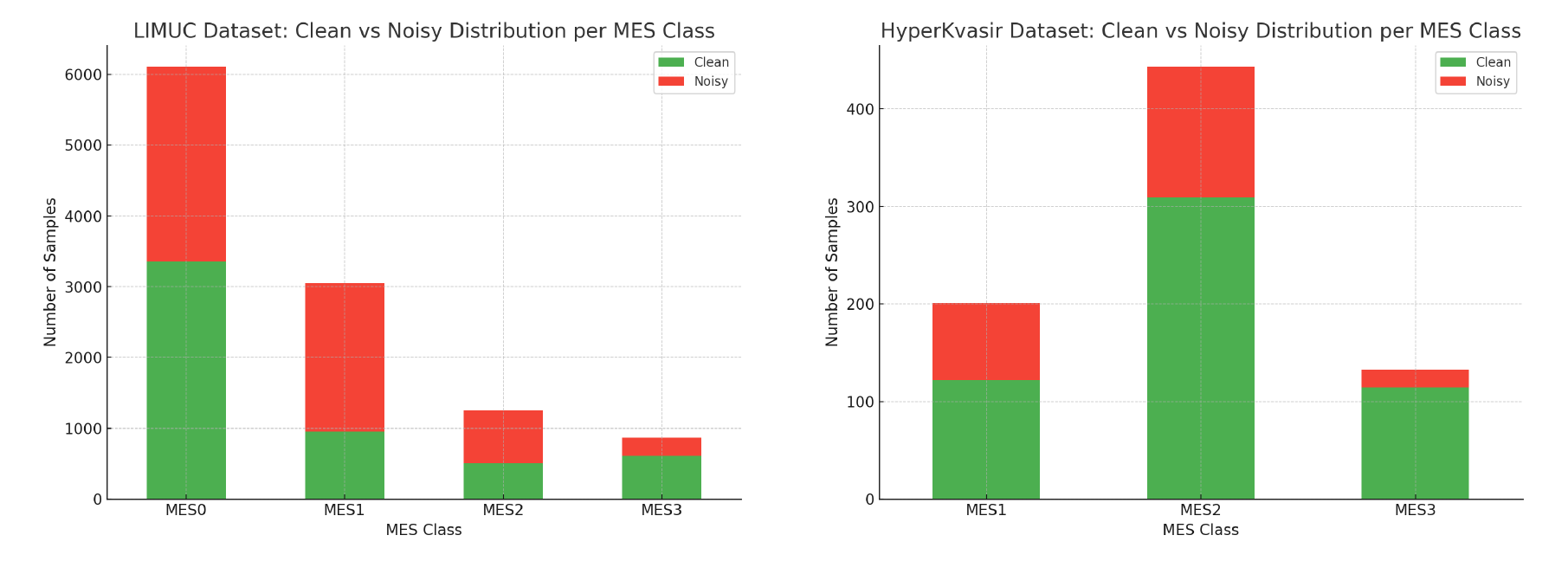}
\caption{Clean vs. noisy sample distribution per MES class across both datasets.}
\label{fig:clean_noisy_distribution}
\end{figure}

Once trained, this quality classifier was applied to all MES-labeled samples in both datasets to generate pseudo-labels (clean or noisy), allowing us to analyze the quality distribution across MES grades. Figure~\ref{fig:clean_noisy_distribution} depicts the resulting quality distribution by MES class. In both datasets, a substantial proportion of noise is observed in the lower MES classes, which complicates the reliable assessment of mild disease activity and further motivates the use of difficulty-aware training strategies.

\begin{table}[H]
\centering
\caption{MES class distribution in LIMUC and HyperKvasir datasets.}
\label{tab:mes_class_distribution_combined}
\begin{tabular}{lcccccc}
\toprule
\multirow{2}{*}{\textbf{MES Class}} & \multicolumn{3}{c}{\textbf{LIMUC}} & \multicolumn{3}{c}{\textbf{HyperKvasir}} \\
\cmidrule(lr){2-4} \cmidrule(lr){5-7}
& Train & Test & Total & Train & Test & Total \\
\midrule
MES-0 & 5,180 & 925 & 6,105 & -- & -- & -- \\
MES-1 & 2,588 & 464 & 3,052 & 161 & 40 & 201 \\
MES-2 & 1,077 & 177 & 1,254 & 355 & 88 & 443 \\
MES-3 & 745   & 120 & 865   & 107 & 26 & 133 \\
\bottomrule
\end{tabular}
\end{table}

\subsection{Implementation Details}
\label{subsec:implementations}

All experiments were conducted on a single NVIDIA GTX 1080 Ti GPU with 11GB of VRAM. All backbone models were initialized with ImageNet-pretrained weights, and training was performed independently for each model. Early stopping was employed based on stagnation in validation accuracy. Batch sizes were adjusted according to GPU memory constraints and empirical training stability.

\begin{table}[H]
\centering
\caption{Training configurations for each backbone architecture.}
\label{tab:training-summary}
\begin{tabular}{llll}
\toprule
\textbf{Model} & \textbf{Optimizer / LR} & \textbf{Scheduler} & \textbf{Batch Size} \\
\midrule
ConvNeXt-Tiny & AdamW (LR=2.5e-4, WD=0.05) & CosineAnnealingLR & 32 \\
ResNet50 (supervised) & SGD (LR=2e-4) & CosineAnnealingLR & 64 \\
ResNet50 (SSL) & LARS (LR=1.6) & CosineAnnealingLR & 64 \\
DenseNet121 & SGD (LR=2e-4, momentum=0.9) & CosineAnnealingLR & 64 \\
EfficientNet-B4 & SGD (LR=0.002) & StepLR (every 40 epochs) & 32 \\
MobileNetV2 & SGD (LR=0.002) & StepLR (every 40 epochs) & 64 \\
ViT-Base (MAE) & AdamW (LR=2.5e-4, WD=0.05) & CosineAnnealingLR & 32 \\
\bottomrule
\end{tabular}
\vspace{2pt}
{\centering
\begin{minipage}{0.9\linewidth}
\centering
{\small\textit{Note.} All pretrained weights were obtained from the MMPretrain library~\cite{mmpretrain}.}
\end{minipage}\par}
\end{table}

%Backbone selection was guided by a balance between model complexity, computational efficiency, and task relevance. ConvNeXt-Tiny and ViT-Base were chosen to evaluate modern CNN and Transformer-based architectures. ResNet50 and DenseNet121 served as strong supervised CNN baselines. SEResNet50, a ResNet variant with attention mechanisms, was also included. MobileNetV2 was used for its lightweight design, enabling deployment in resource-constrained environments. EfficientNet-B4 leveraged Noisy Student weights for better robustness via self-training, and ViT-Base (MAE) was fine-tuned only in its final layers to preserve general-purpose representations.

All supervised baselines and curriculum-based setups are initialized with ImageNet-pretrained weights obtained via standard supervised classification. The final output layer is adjusted to match dataset-specific label spaces: four classes (MES 0–3) for LIMUC and three classes (MES 1–3) for HyperKvasir, since MES 0 labels in HyperKvasir are unreliable. All models are trained with cross-entropy loss unless otherwise stated. The training configurations for each architecture are summarized in Table~\ref{tab:training-summary}.%; although we tested CDW-CE, focal loss, and CORAL, these alternatives did not yield consistent gains and were omitted from the final results.

All models were trained independently under a shared training pipeline, where basic augmentations such as RandomResizedCrop and RandomFlip were consistently applied across all experiments. In addition, we conducted an ablation study (see Section~\ref{subsubsec:augmentation_effect}) to evaluate the impact of three batch-level augmentation techniques: MixUp, CutMix, and ResizeMix. Based on empirical performance, ResizeMix yielded the most consistent improvements in both QWK and accuracy. As a result, it was integrated into both the supervised and CLoE training setups to form the final Supervised + Aug and CLoE configurations.

% aşağıdakiler yukarıda bir yerde belirtilmişti - yeri orası mı bilemiyorum ancak tekrar edilmemesi için bu versiyonu commentledim (Hacer)
%In the case of CLoE, training followed a curriculum learning strategy, where the model progressed from cleaner to noisier data stages based on pseudo-labeled image quality. Phase transitions were triggered when validation accuracy plateaued across several epochs. This staged design enhanced robustness against noise and improved generalization in later phases.

\section{Results and Discussion}
\label{sec:results_discussion}

This section reports the results of our experimental evaluations conducted on the LIMUC and HyperKvasir datasets, focusing on the ordinal classification of MES. Our analysis covers three training paradigms: supervised learning (Supervised and Supervised + Aug), self-supervised learning (SSL), and our proposed curriculum-based strategy (CLoE), which orders examples by estimated reliability and introduces them gradually.

The “+Aug” designation in the tables refers to a combination of standard augmentations such as RandomResizedCrop, RandomFlip, and ResizeMix. As shown in our ablations, including ResizeMix was generally associated with improvements in both accuracy and ordinal agreement (QWK), and it helped preserving transitions between adjacent MES grades. For clarity, all CLoE experiments in this work inherently employ these augmentations, even when “+Aug” is not explicitly indicated.

The remainder of this section is organized as follows. We first analyze the performance of the proposed learning strategies under various backbone architectures. We then compare our results against recent methods from the literature. Lastly, we present ablation studies to isolate the contribution of individual components within our framework.

We report multiple metrics beyond overall accuracy. The macro-averaged F1 score treats all classes equally, offering robustness against class imbalance. The QWK, particularly suited to ordinal classification, captures agreement between predictions and expert annotations while penalizing misclassifications more heavily when they occur between distant classes. Additional metrics, including specificity (true negative rate) and recall (true positive rate), further inform our analysis. For HyperKvasir, we also report micro-averaged F1 to enable direct comparison with prior work.

\subsection{Experimental Analysis}
\label{subsec:experimental_analysis}

We begin by assessing the performance of the proposed CLoE strategy across various backbone architectures, followed by comparisons with baseline supervised and SSL methods. Results for LIMUC are presented in Table~\ref{tab:result_limuc}, while HyperKvasir outcomes are summarized in Table~\ref{tab:result_hyperkvasir}. 

\begin{table}[!htbp]
\caption{
Performance comparison on the LIMUC \cite{0_dataset_gorkem_polat_2022_5827695} dataset across different training strategies and backbone architectures. Bold values indicate the best performance for each metric. All metrics are reported as percentages; QWK is shown on a 0--1 scale.
}

\label{tab:result_limuc}
\begin{tabular}{llllllll}
\hline
\textbf{Method} &
  \textbf{\begin{tabular}[c]{@{}l@{}}Backbone\\ Model\end{tabular}} &
  \textbf{\begin{tabular}[c]{@{}l@{}}Top-1\\ Acc\end{tabular}} &
  \textbf{\begin{tabular}[c]{@{}l@{}}Top-2\\ Acc\end{tabular}} &
  \textbf{\begin{tabular}[c]{@{}l@{}}F1\\ (Macro)\end{tabular}} &
  \textbf{Specificity} &
  \textbf{Recall} &
  \textbf{QWK} \\ \hline
\multirow{4}{*}{Self-Supervised} 
     & SimCLR        & 74.49 & 94.36 & 66.12 & 89.15 & 64.55 & 0.8236 \\
     & SimSiam       & 74.67 & 94.24 & 66.54 & 89.35 & 65.40 & 0.8241 \\
     & MoCoV2        & 75.74 & 94.83 & 67.40 & 90.68 & 66.91 & 0.8352 \\
     & BYOL          & 75.97 & 96.75 & 70.17 & 82.96 & 66.06 & 0.8521 \\ \midrule

\multirow{2}{*}{Supervised}           
 & MobileNetV2     & 75.14 & 97.45 & 68.84 & 90.45 & 69.39 & 0.8452 \\ 
 & ResNet50        & 76.56 & 97.27 & 69.79 & 90.58 & 70.36 & 0.8436 \\ \midrule
 
\multirow{2}{*}{Supervised + Aug}     
 & ResNet50        & 77.40 & 97.50 & 69.98 & 91.74 & 69.12 & 0.8474 \\
 & MobileNetV2     & 78.90 & 97.60 & 72.04 & 91.63 & 71.52 & 0.8696 \\ \midrule

\multirow{7}{*}{CLoE (ours)}      
 & ViT (base p14)  & 76.96 & 97.20 & 70.34 & 90.61 & 67.82 & 0.8432 \\
 & DenseNet121     & 77.63 & 97.51 & 69.60 & 90.87 & 68.30 & 0.8475 \\
 & ResNet50        & 79.10 & 97.72 & 70.14 & 91.12 & 72.63 & 0.8586 \\
 & SEResNet50      & 79.26 & 97.03 & 71.24 & 91.65 & 69.62 & 0.8575 \\
 & EfficientNetB4  & 79.92 & \underline{97.98} & \underline{73.30} & 91.60 & 72.50 & \underline{0.8763} \\
 & MobileNetV2     & \underline{80.03} & 97.80 & 73.12 & \underline{91.63} & \underline{72.80} & 0.8692 \\
 & ConvNeXt (tiny) & \textbf{82.51} & \textbf{98.22} & \textbf{76.79} & \textbf{92.84} & \textbf{76.41} & \textbf{0.8935} \\ \hline
\end{tabular}

\end{table}

\begin{table}[!htbp]
\caption{
Performance comparison on the HyperKvasir \cite{borgli2020hyperkvasir} dataset across different training strategies and backbone architectures. Bold values indicate the best performance for each metric. All metrics are reported as percentages; QWK is shown on a 0--1 scale.
}
\label{tab:result_hyperkvasir}
\begin{tabular}{llllllll}
\hline
\textbf{Method} &
  \textbf{\begin{tabular}[c]{@{}l@{}}Backbone\\ Model\end{tabular}} &
  \textbf{\begin{tabular}[c]{@{}l@{}}Top-1\\ Acc\end{tabular}} &
  \textbf{\begin{tabular}[c]{@{}l@{}}Top-2\\ Acc\end{tabular}} &
  \textbf{\begin{tabular}[c]{@{}l@{}}F1 \\ (Macro)\end{tabular}} &
  \textbf{Specificity} &
  \textbf{Recall} &
  \textbf{QWK} \\ \hline
\multirow{4}{*}{Self-Supervised} 
 & SimCLR  & 74.02 & 96.75 & 66.31 & 81.25 & 62.66 & 0.5674 \\
 & MoCoV2  & 74.02 & 93.50 & 68.23 & 81.89 & 64.02 & 0.6268 \\
 & BYOL    & 75.97 & 96.75 & 70.17 & 82.32 & 66.06 & 0.6542 \\ 
 & SimSiam & 76.62 & 94.80 & 70.65 & 83.56 & 66.44 & 0.6345 \\ \midrule

\multirow{2}{*}{Supervised} 
  & ResNet50        & 72.50 & 96.75 & 68.00 & 80.40 & 65.50 & 0.6121 \\
  & MobileNetV2     & 74.02 & 96.75 & 70.89 & 82.42 & 68.09 & 0.6564 \\ \midrule 

\multirow{2}{*}{Supervised + Aug}     
 & ResNet50        & 73.37 & 96.75 & 69.76 & 81.48 & 64.34 & 0.6263 \\
 & MobileNetV2     & 76.78 & 97.35 & 70.79 & 82.91 & 65.99 & 0.6520 \\ \midrule

\multirow{6}{*}{CLoE (ours)} 
  & DenseNet121     & 68.18 & 96.75 & 55.77 & 75.75 & 54.24 & 0.4496 \\
  & ViT (base p14)  & 72.53 & 96.75 & 73.75 & 78.23 & 63.23 & 0.5235 \\
  & ResNet50        & 74.02 & 96.75 & 69.65 & 81.96 & 67.51 & 0.6934 \\
  & MobileNetV2     & 79.22 & \underline{96.75} & 76.72 & 85.32 & \underline{73.83} & 0.7245 \\
  & EfficientNetB4  & \underline{79.45} & \textbf{97.40} & \underline{76.79} & \textbf{86.78} & \textbf{74.29} & \underline{0.7378} \\
  & ConvNeXt (tiny) & \textbf{80.02} & 96.10 & \textbf{76.85} & \underline{85.92} & 73.31 & \textbf{0.7394} \\ \hline
\end{tabular}

\end{table}

%Among SSL methods, BYOL achieved the best accuracy on the LIMUC dataset, whereas SimSiam led on HyperKvasir. However, the overall performance gap between supervised and SSL methods remained modest across both datasets. This indicates that, in our setting, SSL methods underperform in low-volume, low-variability medical imaging contexts, as reflected in Top-1 Accuracy values where the best SSL models reached 75.97\% on LIMUC and 76.62\% on HyperKvasir, compared to 78.90\% and 76.78\% for the best Supervised+Aug configurations. In particular, contrastive SSL approaches rely on random augmentations to construct positive and negative pairs. In domains like medical imaging, where intra-class variability is naturally low, this strategy can lead to semantically similar instances being placed in opposite classes, ultimately degrading representation quality and reducing ordinal sensitivity. Furthermore, most SSL frameworks do not incorporate inductive biases that reflect ordinal class relationships, limiting their effectiveness for this type of classification task.

Among SSL methods, BYOL achieved the highest accuracy on the LIMUC dataset, whereas SimSiam performed best on HyperKvasir. However, the overall performance gap between supervised and SSL methods was modest across both datasets. In our setting, SSL methods consistently underperformed in low-volume, low-variability medical imaging contexts, as reflected in Top-1 accuracy: the best SSL models reached 75.97\% on LIMUC and 76.62\% on HyperKvasir, compared to 78.90\% and 76.78\% for the best Supervised+Aug configurations. A key limitation of contrastive SSL approaches is their reliance on random augmentations to construct positive and negative pairs. In medical imaging, where intra-class variability is inherently low, this often results in semantically similar instances being treated as negatives, degrading representation quality and reducing ordinal sensitivity. In addition, most SSL frameworks lack inductive biases that capture ordinal class relationships, which further restricts their effectiveness for MES classification.

%Ablation studies (see Section~\ref{subsec:ablation}) show that the use of data augmentation alone, especially ResizeMix, can lead to improvements of up to 2 to 2.5\% in both accuracy and QWK. However, when combined with the structured learning framework provided by CLoE, performance gains exceed 5\%. The results indicate that CLoE provides benefits beyond augmentation, potentially introducing a degree of structured learning. In our experiments, the CLoE setting focused on more reliable training samples while maintaining ordinal consistency in the latent space, which may help mitigate the influence of noisy labels.
Ablation studies (see Section~\ref{subsec:ablation}) show that data augmentation alone, particularly ResizeMix, improves performance by up to 2–2.5\% in both accuracy and QWK. When combined with the structured training schedule of CLoE, gains exceed 5\%, demonstrating that CLoE offers benefits beyond augmentation. By prioritizing reliable samples and preserving ordinal consistency in the latent space, CLoE helps reducing the impact of noisy labels and introduces a form of structured learning that is especially effective in this setting.

Across both datasets, the ConvNeXt (tiny) backbone consistently achieved the best results under the proposed CLoE configuration, reaching 82.51\% accuracy and 0.8935 QWK on LIMUC, and 80.02\% accuracy and 0.7394 QWK on HyperKvasir. ConvNeXt integrates modern CNN design elements such as LayerNorm, GELU activations, and large convolutional kernels, alongside ViT-inspired principles. This allows it to effectively capture both local and global patterns. Its patch-based representation and progressive downsampling may support long-range dependency modeling, which could be beneficial for preserving smooth transitions between adjacent ordinal categories.

\begin{figure}[H]
\centering
\includegraphics[width=\linewidth]{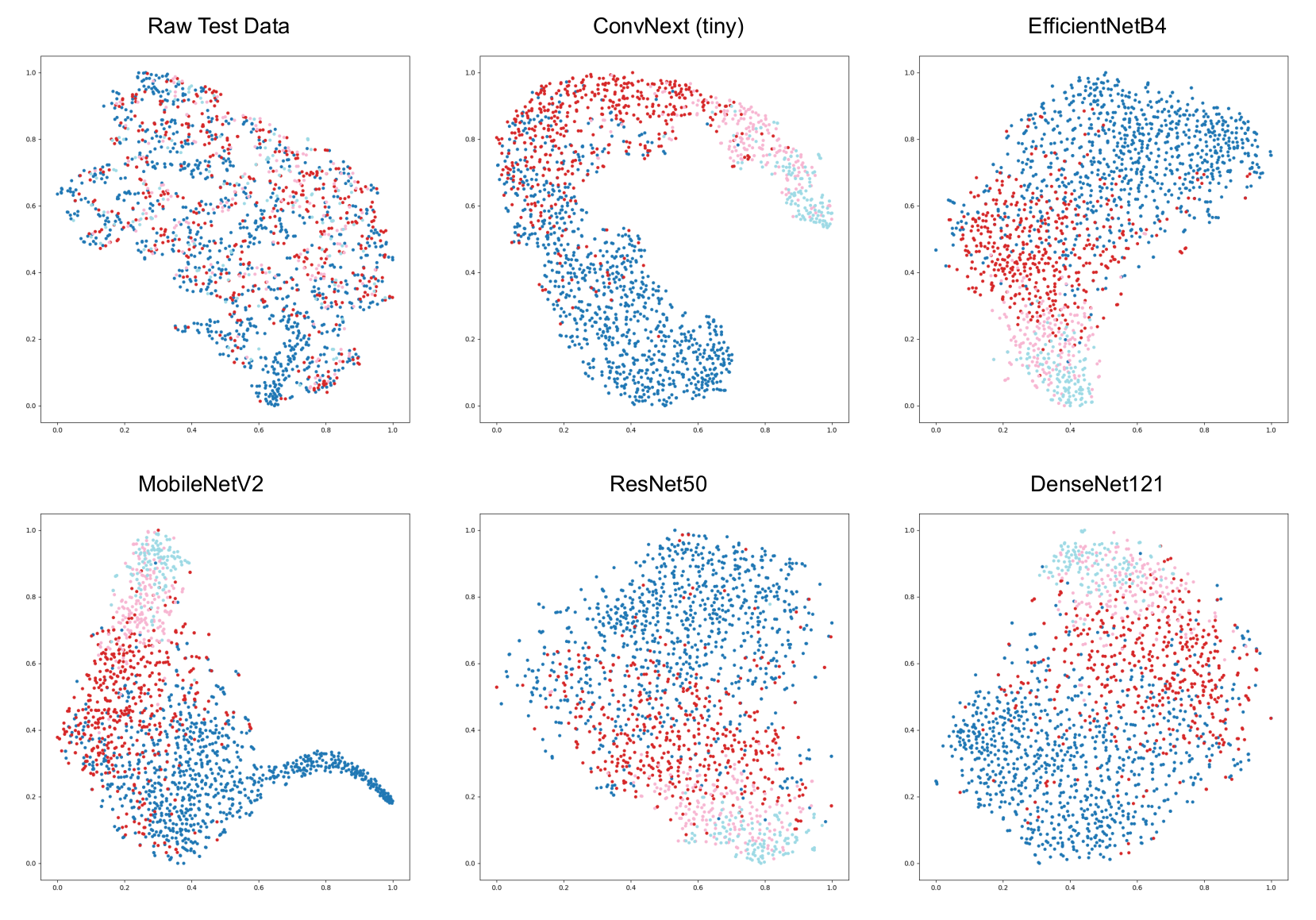}
\caption{%
Two-dimensional visualization of feature vectors from different backbone architectures trained on the LIMUC dataset using the CLoE strategy, obtained via t-SNE (perplexity=30, learning rate=200, 1000 iterations) on the test set. %This visualization is qualitative; quantitative comparisons rely on the metrics reported in Table~\ref{tab:result_limuc}.%
}
\label{fig:limuc_tsne}
\end{figure}

Figure~\ref{fig:limuc_tsne} provides a two-dimensional t-SNE visualization of learned feature representations for different backbones on the LIMUC dataset under the CLoE setting. Among the evaluated architectures, ConvNeXt produced the clearest class separation, with smooth transitions only between adjacent categories. While EfficientNet-B4 and MobileNetV2 also preserved distinguishable class clusters, their boundaries were slightly more diffused. In contrast, classical architectures such as ResNet50 and DenseNet121 showed more overlap between neighboring classes in this qualitative t-SNE view, which may indicate weaker ordinal separation but does not constitute statistical evidence. We therefore treat t-SNE as a visual aid and rely on quantitative metrics, particularly QWK and Macro-F1 in Tables~\ref{tab:result_limuc}, to assess ordinal behavior.

An additional observation is that lightweight models like MobileNetV2, despite having only 3.5M parameters and 0.32 GFLOPs, reach competitive accuracy and high QWK under the CLoE configuration (see Tables~\ref{tab:result_limuc}–\ref{tab:result_hyperkvasir}). Similarly, compact models such as EfficientNet-B4 and SEResNet50 also perform strongly. These findings suggest that even lightweight models, when combined with the proposed strategy, can reach strong performance. Thus, model capacity alone does not fully explain the observed results.

Findings from the HyperKvasir dataset are largely consistent with those observed on LIMUC. Once again, ConvNeXt (tiny) achieved the highest overall performance, while both MobileNetV2 and EfficientNet-B4 surpassed the 79\% accuracy threshold. Although SimSiam outperformed other SSL variants, models trained with CLoE consistently yielded superior results. These outcomes are consistent with our LIMUC findings and suggest robustness across the two datasets considered. Note that LIMUC is larger and more homogeneous, whereas HyperKvasir shows greater variability in anatomical content and acquisition conditions. Such diversity can be challenging for SSL methods; in our experiments, CLoE performed comparatively well, which may be linked to its emphasis on higher-confidence samples and to mechanisms that encourage ordinal structure.

\begin{table}[h]
\centering
\caption{Parameter and FLOPs values of the backbone models used in this study.}
\label{tab:backbone_params_flops}
\begin{tabular}{lll}
\hline
\textbf{Model} & \textbf{Params (M)} & \textbf{FLOPs (G)} \\
\hline
SimCLR (ResNet50) & 27.97 & 4.11 \\
SimSiam (ResNet50) & 38.20 & 4.11 \\
MoCoV2 (ResNet50) & 55.93 & 4.11 \\
BYOL (ResNet50) & 68.02 & 4.11 \\  \midrule

MobileNetV2 & 3.50 & 0.32 \\
DenseNet121 & 7.98 & 2.88 \\
EfficientNet-B4 & 19.34 & 4.66 \\
ResNet50 & 25.56 & 4.11 \\ 
SEResNet50 & 28.09 & 4.13 \\
ConvNeXt (tiny) & 28.59 & 4.66 \\
ViT (base p14) & 86.57 & 17.58 \\
\hline
\end{tabular}

\end{table}

Finally, the proposed strategy achieved competitive predictive performance while maintaining computational efficiency. As summarized in Table~\ref{tab:backbone_params_flops}, backbones such as ConvNeXt (tiny), MobileNetV2, and EfficientNet-B4 strike a favorable balance between accuracy and resource usage. In our evaluation, ConvNeXt (tiny) obtained the best results among the backbones we considered (28.59M parameters, 4.66 GFLOPs) and performed better than ViT (base) on both accuracy and efficiency. As shown in Table 5, the FLOPs of SSL methods are essentially identical to that of ResNet-50, since their projection heads are typically lightweight MLPs that increase parameter counts while adding negligible computational overhead.

\subsection{State of the Art Comparison}
\label{subsec:sota_comparison}

Previous studies on the LIMUC and HyperKvasir datasets have largely focused on supervised or SSL learning paradigms, typically evaluating performance using general metrics such as accuracy, macro-F1, or occasionally QWK. However, as shown in Table~\ref{tab:sota_limuc} and Table~\ref{tab:sota_hyperkvasir}, many approaches do not explicitly handle two common dataset properties: ordered (ordinal) labels and the presence of noisy annotations.

Polat et al.\cite{1_polat2022class} proposed an ordinal-regression–style objective that treats class indices as ordered and uses a class distance–weighted cross-entropy, using InceptionV3, MobileNetV3, and DenseNet121, reporting up to 69\% macro-F1 and 0.85 QWK. A subsequent study\cite{3_polat2023improving} extended this framework to ViT-based architectures, achieving similar results with approximately 68\% macro-F1 and 0.84 QWK. Although these models implicitly incorporated ordinal information through the loss function, they did not explicitly down-weight uncertain or noisy labels.

Another study, Çağlar et al.~\cite{2_ccauglar2023ulcerative} attempted to mitigate class imbalance by combining GAN-based data augmentation with active sample selection. While they reached 67\% macro-F1 and 0.809 QWK, ordinal relationships were considered mainly in data selection, not directly encoded in the learning objective.

Xu et al.~\cite{5_xu2024ssl} introduced a contrastive self-supervised method named SSL-CPCD, based on SimCLR and a ResNet50 backbone, attaining 75.9\% accuracy. In settings with low intra-class variability, contrastive learning can be sensitive to false negatives; this method also did not model ordinal alignment or sample reliability.

In a more recent study, Polat et al.\cite{9_polat2025class} used MixUp-based data augmentation, achieving 77.5\% accuracy. While effective in many tasks, MixUp’s linear interpolation can blur boundaries between ordered categories. In our ablations (Table\ref{tab:ablation_augmentation}), ResizeMix led to more consistent improvements for the ordinal setting.

The EndoDINO framework~\cite{dermyer2025endodino} introduced a pretrained model based on DINOv2, trained on endoscopic video frames from the HyperKvasir dataset. Across ViT variants (including ViT-g/14), the best reported macro-F1 was 74.8\%. In our setup, CLoE with a compact ConvNeXt-tiny backbone achieved 76.79\% macro-F1 and 0.893 QWK on LIMUC.

Overall, our results indicate that CLoE addresses two practical needs: (i) explicit ordinal alignment to respect the order of adjacent classes and (ii) a way to reduce the influence of uncertain labels. Augmentations that respect label order, such as ResizeMix, serve as a simple alternative to generic policies in this setting.

Empirical results support the approach. On the LIMUC dataset, CLoE achieved 82.51\% accuracy, 76.79\% macro-F1, and 0.893 QWK. On the more heterogeneous HyperKvasir dataset, performance reached 80.02\% accuracy, 74.02\% macro-F1, and 0.739 QWK. These scores exceed those reported for SSL-CPCD~\cite{5_xu2024ssl} and EndoDINO~\cite{dermyer2025endodino} under their respective setups. All results were obtained with computationally efficient backbones.

\begin{table}[H]
\caption{
Comparison of CLoE + Aug with SOTA on the LIMUC dataset. Bold values indicate the best overall performance. All metrics are reported as percentages; QWK is shown on a 0--1 scale.}
\label{tab:sota_limuc}
\begin{tabular}{llllllll}
\hline
\textbf{Model/Method} &
  \textbf{\begin{tabular}[c]{@{}l@{}}Backbone\\ Model\end{tabular}} &
  \textbf{\begin{tabular}[c]{@{}l@{}}Top-1\\ Acc\end{tabular}} &
  \textbf{\begin{tabular}[c]{@{}l@{}}Top-2\\ Acc\end{tabular}} &
  \textbf{\begin{tabular}[c]{@{}l@{}}F1\\ (Macro)\end{tabular}} &
  \textbf{Specificity} &
  \textbf{Recall} &
  \textbf{QWK} \\ \hline
  \multirow{3}{*}{Supervised + Reg. \cite{3_polat2023improving}} 
 & InceptionV3        & --          & -- & 68.8 & -- & -- & 0.852 \\
 & MobileNetV3 (large)   & --          & -- & 69.5 & -- & -- & 0.847 \\
 & DenseNet121         & --          & -- & 69.7 & -- & -- & 0.854 \\ \hline

SimCLR+DCL \cite{5_xu2024ssl}                 & ResNet50 + Att. & 75.80          & 93.67 & 67.55 & 86.69 & 69.52 & 0.8287 \\

MoCoV2+CLD \cite{5_xu2024ssl}                 & ResNet50 + Att. & 75.98          & 95.36 & 68.12 & 87.09 & 70.47 & 0.8333 \\

PIRL \cite{5_xu2024ssl}                       & ResNet50 + Att. & 77.4           & 96.1  & 69.18 & 88.93 & 71.33 & 0.8460 \\

PIRL+PLD \cite{5_xu2024ssl}                   & ResNet50 + Att. & 78.47          & 97.07 & 71.67 & 89.33 & 71.46 & 0.8563 \\
SSL-CPCD  \cite{5_xu2024ssl}                  & ResNet50 + Att. & 79.77          & 97.5  & 72.79 & 90.08 & 72.59 & 0.8746 \\ \midrule
EndoDINO  \cite{dermyer2025endodino}          & ViT (base p14)  & --             & --    & 70.6 & --    & --    & --    \\
EndoDINO  \cite{dermyer2025endodino}          & ViT (giant p14) & --             & --    & 71.5 & --    & --    & --    \\ \midrule
\multirow{3}{*}{CLoE (ours)}             
 & EfficientNetB4  & 79.92 & \underline{97.98} & \underline{73.3} & 91.6 & 72.5  & \underline{0.8763} \\
 & MobileNetV2     & \underline{80.03} & 97.8  & 73.12 & \underline{91.63} & \underline{72.8}  & 0.8692 \\
 & ConvNeXt (tiny) & \textbf{82.51} & \textbf{98.22} & \textbf{76.79} & \textbf{92.84} & \textbf{76.41} & \textbf{0.8935} \\ \hline
\end{tabular}

\end{table}

\begin{table}[H]
\caption{Comparison of CLoE + Aug with SOTA on the HyperKvasir dataset. Bold values indicate the best overall performance. All metrics are reported as percentages; QWK is shown on a 0--1 scale.}
\label{tab:sota_hyperkvasir}
\begin{tabular}{lllllllll}
\hline
\textbf{Model / Method} &
  \textbf{\begin{tabular}[c]{@{}l@{}}Backbone\\  Model\end{tabular}} &
  \textbf{\begin{tabular}[c]{@{}l@{}}Top-1\\ Acc\end{tabular}} &
  \textbf{\begin{tabular}[c]{@{}l@{}}Top-2\\ Acc\end{tabular}} &
  \textbf{\begin{tabular}[c]{@{}l@{}}F1 \\ (Macro)\end{tabular}} &
  \textbf{\begin{tabular}[c]{@{}l@{}}F1 \\ (Micro)\end{tabular}} &
  \textbf{Specificity} &
  \textbf{Recall} &
  \textbf{QWK} \\ \hline
\multirow{3}{*}{EndoDINO \cite{dermyer2025endodino}}          
  & ViT (base p14)  & --             & --    & 74.0  & 77.6  & --    & --    & --    \\
  & ViT (large p14) & --             & --    & 74.0  & 76.8  & --    & --    & --    \\
  & ViT (giant p14) & --             & --    & 74.8  & 77.9  & --    & --    & --    \\ \midrule
\multirow{3}{*}{CLoE (ours)} 
  & MobileNetV2     & 79.22 & \underline{96.75} & 76.72 & 79.06 & 85.32 & \underline{73.83} & 0.7245 \\
  & EfficientNetB4  & \underline{79.45} & \textbf{97.4} & \underline{76.79} & \underline{79.12} & \underline{86.78} & \textbf{74.29} & \underline{0.7378} \\
  & ConvNeXt (tiny) & \textbf{80.02} & 96.1 & \textbf{76.85} & \textbf{79.69} & \textbf{85.92} & 73.31 & \textbf{0.7394} \\ \hline
\end{tabular}

\end{table}

\subsection{Ablation Study}
\label{subsec:ablation}

This section presents the ablation study conducted to evaluate the effectiveness of the proposed CLoE strategy. The experiments were performed on LIMUC, using two representative backbone architectures: MobileNetV2 and ResNet50. The evaluation was structured around three distinct training protocols: STD-A, which refers to standard training using all data including noisy labels; STD-C, which uses only clean-labeled data; and CL, a curriculum learning strategy in which training samples are introduced progressively according to their estimated reliability. This curriculum progression begins with high-confidence samples labeled as clean, followed by moderately reliable samples categorized as mixed, and concludes with low-confidence examples considered noisy. Reliability scores are computed from pseudo-label confidence/consistency on the training split (Sec.~\ref{sec:datasets_and_implementations}). Importantly, all results reported in this ablation section are based on validation performance, while the final test results are provided in (Sec.~\ref{subsec:experimental_analysis}). The LIMUC validation set comprises 960 samples in total, distributed as MES-0: 516, MES-1: 254, MES-2: 111, and MES-3: 79.

\subsubsection{Effect of Data Augmentation}
\label{subsubsec:augmentation_effect}

To further explore the combined effect of curriculum learning and data augmentation, we evaluated three batch-based strategies: MixUp, CutMix, and ResizeMix. These approaches are widely used in medical image classification \cite{singh2021metamed, ozdemir2024meta}. On LIMUC (Table~\ref{tab:ablation_augmentation}), most augmentations improved performance compared to the baseline. The largest gains were observed when combining curriculum learning with ResizeMix, reaching 81.35\% accuracy and 0.882 QWK with the MobileNetV2 backbone (and 78.75\% accuracy, 0.854 QWK with ResNet50). 

ResizeMix creates synthetic training samples by resizing a random region from one image and pasting it onto another. While effective, its behavior may differ depending on the training protocol. Under STD-A, where clean and noisy samples are mixed indiscriminately, ResizeMix could potentially propagate label noise by combining unreliable examples with clean ones. In contrast, curriculum learning introduces subsets of data progressively: in the early phases, only clean samples are available, so ResizeMix operates primarily between more reliable examples. This may help the model to focus on stable features at the beginning of training, which could in turn contribute to more robust representations once noisier samples are incorporated. We hypothesize that this interaction explains, at least in part, why ResizeMix and CL together yield stronger performance than either alone.

\begin{table}[H]
\centering

\caption{Performance comparison on the LIMUC validation set using different augmentation methods with and without CL. Best results for each configuration group are in bold.}

\label{tab:ablation_augmentation}
\begin{tabular}{llllllll}
\hline
\multirow{2}{*}{\textbf{Dataset}} & \multirow{2}{*}{\textbf{\begin{tabular}[c]{@{}l@{}}Backbone\\ Model\end{tabular}}} & \multirow{2}{*}{\textbf{\begin{tabular}[c]{@{}l@{}}Training\\  Protocol\end{tabular}}} & \multirow{2}{*}{\textbf{Acc}} & \multirow{2}{*}{\textbf{F1}} & \multirow{2}{*}{\textbf{QWK}} & \multirow{2}{*}{\textbf{Precision}} & \multirow{2}{*}{\textbf{Recall}} \\
 &  &  &  &  &  &  &  \\ \hline
\multirow{8}{*}{\textbf{LIMUC}} & \multirow{5}{*}{MobileNetV2} 
    & STD-A                  & 78.54 & 71.03 & 0.8645 & 73.87 & 69.83 \\
 &  & STD-A + CutMix         & 78.87 & 70.17 & 0.8545 & 74.80 & 68.08 \\
 &  & STD-A + MixUp          & 78.95 & 71.35 & 0.8781 & 72.95 & 70.67 \\
 &  & STD-A + ResizeMix      & 79.79 & 72.82 & 0.8805 & 74.61 & 72.16 \\
 &  & CL + ResizeMix (CLoE)  & \textbf{81.35} & \textbf{75.20} & \textbf{0.8828} & \textbf{76.98} & \textbf{73.71} \\ \cline{2-8}
 & \multirow{3}{*}{ResNet50} 
    & STD-A                  & 76.97 & 67.57 & 0.8419 & 71.80 & 65.13 \\
 &  & STD-A + ResizeMix      & 77.87 & 69.92 & 0.8530 & 72.70 & 67.82 \\
 &  & CL + ResizeMix (CLoE)  & \textbf{78.75} & \textbf{70.10} & \textbf{0.8540} & \textbf{73.90} & \textbf{69.74} \\ \hline
\end{tabular}

\end{table}

\subsubsection{Comparison of Training Protocols}
\label{subsubsec:curriculum_effect}

As shown in Table~\ref{tab:ablation_clean_noisy}, the CL protocol generally achieves the best results across both backbones. Compared to STD-A, which suffers from the presence of noisy labels, CL yields modest but consistent gains (about 1–2\% across metrics). STD-C, while benefiting from clean labels, is limited by the reduced training set size and tends to underperform on the full validation set. This suggests that by progressively incorporating data with different reliability levels, CL may strike a balance between robustness to label noise and sufficient data coverage. Although the improvements are not large in absolute terms, the consistency across backbones suggests that curriculum learning provides a more stable training dynamic than either standard protocol alone.

\begin{table}[H]
\centering
\caption{Comparison of training protocols (STD-A, STD-C, CL) on the LIMUC \cite{0_dataset_gorkem_polat_2022_5827695} dataset using MobileNetV2 \cite{sandler2018mobilenetv2} and ResNet50 \cite{resnet_targ2016}. Bold values indicate the best result within each group. All metrics are reported as percentages; QWK is scaled to the 0--1 range.}
\label{tab:ablation_clean_noisy}
\begin{tabular}{llllllll}
\hline
\multirow{2}{*}{\textbf{Dataset}} & 
\multirow{2}{*}{\textbf{\begin{tabular}[c]{@{}l@{}}Backbone\\  Model\end{tabular}}} & 
\multirow{2}{*}{\textbf{\begin{tabular}[c]{@{}l@{}}Training\\  Protocol\end{tabular}}} & 
\multirow{2}{*}{\textbf{Acc}} & 
\multirow{2}{*}{\textbf{F1}} & 
\multirow{2}{*}{\textbf{QWK}} & 
\multirow{2}{*}{\textbf{Precision}} & 
\multirow{2}{*}{\textbf{Recall}} \\
 &  &  &  &  &  &  &  \\ \hline

\multirow{6}{*}{\textbf{LIMUC}} 
 & \multirow{3}{*}{MobileNetV2} 
 & STD-C & 76.66 & 68.25 & 0.8547 & 69.78 & 69.01 \\
 &       & STD-A & 78.54 & 71.03 & 0.8645 & 73.87 & 69.83 \\
 &       & CL    & \textbf{79.56} & \textbf{73.64} & \textbf{0.8683} & \textbf{76.18} & \textbf{71.78} \\ \cline{2-8} 
 & \multirow{3}{*}{ResNet50} 
 & STD-C & 76.87 & 69.06 & 0.8584 & 70.69 & 67.97 \\
 &       & STD-A & 76.97 & 67.57 & 0.8419 & 71.80 & 65.13 \\
 &       & CL    & \textbf{77.87} & \textbf{70.02} & \textbf{0.8652} & \textbf{72.95} & \textbf{68.35} \\ \hline

\end{tabular}

\end{table}

\section{Conclusion}
\label{sec:conclusion}

In this work, we proposed CLoE, a difficulty-aware training framework designed for ordinal classification tasks in medical imaging, with a specific focus on MES grading in ulcerative colitis. Compared with prior work, CLoE explicitly targets two fundamental challenges commonly encountered in real-world clinical datasets: (i) the presence of noisy annotations caused by inter-observer variability and poor image quality, and (ii) the ordinal nature of disease severity scores.

Our method leverages image quality as a proxy for annotation reliability, obtained via a lightweight classifier trained on BBPS labels. These difficulty estimates enable a curriculum learning strategy that introduces training data in a structured order, beginning with clean samples and gradually incorporating more uncertain examples. In parallel, the use of ResizeMix augmentation helps maintain semantic consistency across adjacent MES classes and can improve generalization while preserving ordinal structure.

We validated the effectiveness of CLoE across two endoscopic datasets, namely LIMUC and HyperKvasir, and six different backbone families. Our results generally showed improvements in both accuracy and QWK. The method outperformed strong supervised and self-supervised baselines, including large-scale ViT models and contrastive SSL approaches, in our experimental setup. Even when implemented with lightweight architectures such as ConvNeXt-Tiny, the method achieved 82.51\% accuracy and 0.893 QWK on LIMUC. These results suggest that it is possible to combine ordinal consistency with computational efficiency.

Ablation studies further indicated that combining curriculum-based scheduling with data augmentation provides additive benefits. Curriculum learning is associated with improved robustness to label noise, and ResizeMix helps preserve class structure during mixing. Importantly, the proposed method, which works across diverse backbones, is computationally efficient, and does not require changes to model architectures. This makes it a practical candidate for deployment in resource-constrained or clinical settings.

Future work could explore the application of CLoE to other ordinal classification tasks, its extension to multi-modal inputs such as pathology reports or clinical metadata, and the integration of uncertainty estimation techniques to improve reliability modeling and better handle ambiguous cases.

%Bibliography
\bibliographystyle{unsrt}  
\bibliography{references}  

\begin{thebibliography}{10}

\bibitem{feuerstein2014ulcerative}
Joseph~D Feuerstein and Adam~S Cheifetz.
\newblock Ulcerative colitis: epidemiology, diagnosis, and management.
\newblock In {\em Mayo Clinic Proceedings}, volume~89, pages 1553--1563. Elsevier, 2014.

\bibitem{0_dataset_gorkem_polat_2022_5827695}
Gorkem Polat, Haluk~Tarik Kani, Ilkay Ergenc, Yesim~Ozen Alahdab, Alptekin Temizel, and Ozlen Atug.
\newblock {Labeled Images for Ulcerative Colitis (LIMUC) Dataset}, March 2022.

\bibitem{3_polat2023improving}
Gorkem Polat, Haluk~Tarik Kani, Ilkay Ergenc, Yesim Ozen~Alahdab, Alptekin Temizel, and Ozlen Atug.
\newblock Improving the computer-aided estimation of ulcerative colitis severity according to mayo endoscopic score by using regression-based deep learning.
\newblock {\em Inflammatory Bowel Diseases}, 29(9):1431--1439, 2023.

\bibitem{9_polat2025class}
Gorkem Polat, {\"U}mit~Mert {\c{C}}a{\u{g}}lar, and Alptekin Temizel.
\newblock Class distance weighted cross entropy loss for classification of disease severity.
\newblock {\em Expert Systems with Applications}, 269:126372, 2025.

\bibitem{11_lee2025artificial}
Michelle Chae~Min Lee, Armin Farahvash, and Petros Zezos.
\newblock Artificial intelligence for classification of endoscopic severity of inflammatory bowel disease: a systematic review and critical appraisal.
\newblock {\em Inflammatory Bowel Diseases}, page izaf050, 2025.

\bibitem{1_polat2022class}
Gorkem Polat, Ilkay Ergenc, Haluk~Tarik Kani, Yesim~Ozen Alahdab, Ozlen Atug, and Alptekin Temizel.
\newblock Class distance weighted cross-entropy loss for ulcerative colitis severity estimation.
\newblock In {\em Annual Conference on Medical Image Understanding and Analysis}, pages 157--171. Springer, 2022.

\bibitem{7_kadota2024deep}
Takeaki Kadota, Hideaki Hayashi, Ryoma Bise, Kiyohito Tanaka, and Seiichi Uchida.
\newblock Deep bayesian active learning-to-rank with relative annotation for estimation of ulcerative colitis severity.
\newblock {\em Medical Image Analysis}, 97:103262, 2024.

\bibitem{13_lee2025multi}
Jaehyuk Lee and Eunchan Kim.
\newblock Multi-task deep learning framework for enhancing mayo endoscopic score classification in ulcerative colitis.
\newblock {\em Digital Health}, 11:20552076251356396, 2025.

\bibitem{5_xu2024ssl}
Ziang Xu, Jens Rittscher, and Sharib Ali.
\newblock Ssl-cpcd: Self-supervised learning with composite pretext-class discrimination for improved generalisability in endoscopic image analysis.
\newblock {\em IEEE Transactions on Medical Imaging}, 2024.

\bibitem{10_margapuri2025diagnosis}
Venkat Margapuri.
\newblock Diagnosis and severity assessment of ulcerative colitis using self supervised learning.
\newblock In {\em 2025 IEEE Symposium on Computational Intelligence in Health and Medicine Companion (CIHM Companion)}, pages 1--5. IEEE, 2025.

\bibitem{jimenez2022curriculum}
Amelia Jim{\'e}nez-S{\'a}nchez, Diana Mateus, Sonja Kirchhoff, Chlodwig Kirchhoff, Peter Biberthaler, Nassir Navab, Miguel A~Gonz{\'a}lez Ballester, and Gemma Piella.
\newblock Curriculum learning for improved femur fracture classification: Scheduling data with prior knowledge and uncertainty.
\newblock {\em Medical Image Analysis}, 75:102273, 2022.

\bibitem{ma2024adaptive}
Siteng Ma, Honghui Du, Kathleen~M Curran, Aonghus Lawlor, and Ruihai Dong.
\newblock Adaptive curriculum query strategy for active learning in medical image classification.
\newblock In {\em International Conference on Medical Image Computing and Computer-Assisted Intervention}, pages 48--57. Springer, 2024.

\bibitem{abbas2025clog}
Asmaa Abbas, Mohamed~Medhat Gaber, and Mohammed~M Abdelsamea.
\newblock Clog-cd: Curriculum learning based on oscillating granularity of class decomposed medical image classification.
\newblock {\em IEEE Transactions on Emerging Topics in Computing}, 2025.
\newblock Early Access.

\bibitem{burduja2021unsupervised}
Mihail Burduja and Radu~Tudor Ionescu.
\newblock Unsupervised medical image alignment with curriculum learning.
\newblock In {\em 2021 IEEE International Conference on Image Processing (ICIP)}, pages 3787--3791. IEEE, 2021.

\bibitem{borgli2020hyperkvasir}
Hanna Borgli, Vajira Thambawita, Pia~H Smedsrud, Steven Hicks, Debesh Jha, Sigrun~L Eskeland, Kristin~Ranheim Randel, Konstantin Pogorelov, Mathias Lux, Duc Tien~Dang Nguyen, et~al.
\newblock Hyperkvasir, a comprehensive multi-class image and video dataset for gastrointestinal endoscopy.
\newblock {\em Scientific data}, 7(1):283, 2020.

\bibitem{bengio2009curriculum}
Yoshua Bengio, J{\'e}r{\^o}me Louradour, Ronan Collobert, and Jason Weston.
\newblock Curriculum learning.
\newblock In {\em Proceedings of the 26th Annual International Conference on Machine Learning}, pages 41--48. ACM, 2009.

\bibitem{hacohen2019power}
Guy Hacohen and Daphna Weinshall.
\newblock On the power of curriculum learning in training deep networks.
\newblock {\em International Conference on Machine Learning (ICML)}, pages 2535--2544, 2019.

\bibitem{resnet_targ2016}
Sasha Targ, Diogo Almeida, and Kevin Lyman.
\newblock Resnet in resnet: Generalizing residual architectures.
\newblock {\em arXiv preprint arXiv:1603.08029}, 2016.

\bibitem{densenet_huang2017densely}
Gao Huang, Zhuang Liu, Laurens Van Der~Maaten, and Kilian~Q Weinberger.
\newblock Densely connected convolutional networks.
\newblock In {\em Proceedings of the IEEE conference on computer vision and pattern recognition}, pages 4700--4708, 2017.

\bibitem{sandler2018mobilenetv2}
Mark Sandler, Andrew Howard, Menglong Zhu, Andrey Zhmoginov, and Liang-Chieh Chen.
\newblock Mobilenetv2: Inverted residuals and linear bottlenecks.
\newblock In {\em Proceedings of the IEEE conference on computer vision and pattern recognition}, pages 4510--4520, 2018.

\bibitem{convnext_liu2022convnet}
Zhuang Liu, Hanzi Mao, Chao-Yuan Wu, Christoph Feichtenhofer, Trevor Darrell, and Saining Xie.
\newblock A convnet for the 2020s.
\newblock In {\em Proceedings of the IEEE/CVF conference on computer vision and pattern recognition}, pages 11976--11986, 2022.

\bibitem{efficientnet_tan2019}
Mingxing Tan and Quoc Le.
\newblock Efficientnet: Rethinking model scaling for convolutional neural networks.
\newblock In {\em International conference on machine learning}, pages 6105--6114. PMLR, 2019.

\bibitem{vit_dosovitskiy2020image}
Alexey Dosovitskiy, Lucas Beyer, Alexander Kolesnikov, Dirk Weissenborn, Xiaohua Zhai, Thomas Unterthiner, Mostafa Dehghani, Matthias Minderer, Georg Heigold, Sylvain Gelly, et~al.
\newblock An image is worth 16x16 words: Transformers for image recognition at scale.
\newblock {\em arXiv preprint arXiv:2010.11929}, 2020.

\bibitem{4_li2023comparative}
Chenxi Li, Jiawei Yang, Yuxin Qin, Lulu Lv, and Tao Li.
\newblock A comparative study of resnet and densenet in the diagnosis of colitis severity.
\newblock In {\em International Workshop on Internet of Things of Big Data for Healthcare}, pages 102--110. Springer, 2023.

\bibitem{6_al2024ulcerative}
Mohammed Al-Refai, Shahed Alkhaza’leh, and Ahmad Alzu’bi.
\newblock Ulcerative colitis image classification using federated deep learning.
\newblock In {\em International Conference on Intelligent Systems and Pattern Recognition}, pages 207--219. Springer, 2024.

\bibitem{15_shiku2025ordinal}
Kaito Shiku, Kazuya Nishimura, Daiki Suehiro, Kiyohito Tanaka, and Ryoma Bise.
\newblock Ordinal multiple-instance learning for ulcerative colitis severity estimation with selective aggregated transformer.
\newblock In {\em 2025 IEEE/CVF Winter Conference on Applications of Computer Vision (WACV)}, pages 4290--4299. IEEE, 2025.

\bibitem{12_gutierrez2025ulcerative}
Benjamin Gutierrez-Becker, Stefan Fraessle, Heming Yao, Jerome Luscher, Rafal Girycki, Bartosz Machura, Janusz Czornik, Jaroslaw Goslinsky, Marek Pitura, Steven Levitte, et~al.
\newblock Ulcerative colitis severity classification and localized extent (uc-scale): An artificial intelligence scoring system for a spatial assessment of disease severity in ulcerative colitis.
\newblock {\em Journal of Crohn's and Colitis}, 19(1):jjae187, 2025.

\bibitem{14_zhang2025afr}
Kun Zhang, Qianru Yu, Yansheng Liu, Yumeng Duan, Yingying Lou, and Weichao Xu.
\newblock Afr: An image-aided diagnostic approach for ulcerative colitis.
\newblock {\em Biomedical Signal Processing and Control}, 105:107542, 2025.

\bibitem{17_sano2025explainable}
Masaya Sano, Yasuhiro Kanatani, Takashi Ueda, Shota Nemoto, Yurin Miyake, Naoko Tomita, and Hidekazu Suzuki.
\newblock Explainable artificial intelligence for prediction of refractory ulcerative colitis: analysis of a japanese nationwide registry.
\newblock {\em Annals of Medicine}, 57(1):2499960, 2025.

\bibitem{simclr_chen2020simple}
Ting Chen, Simon Kornblith, Mohammad Norouzi, and Geoffrey Hinton.
\newblock A simple framework for contrastive learning of visual representations.
\newblock In {\em International conference on machine learning}, pages 1597--1607. PmLR, 2020.

\bibitem{mocov2_chen2003improved}
Xinlei Chen, Haoqi Fan, Ross Girshick, and Kaiming He.
\newblock Improved baselines with momentum contrastive learning. arxiv 2020.
\newblock {\em arXiv preprint arXiv:2003.04297}, 2003.

\bibitem{byol_grill2020bootstrap}
Jean-Bastien Grill, Florian Strub, Florent Altch{\'e}, Corentin Tallec, Pierre Richemond, Elena Buchatskaya, Carl Doersch, Bernardo Avila~Pires, Zhaohan Guo, Mohammad Gheshlaghi~Azar, et~al.
\newblock Bootstrap your own latent-a new approach to self-supervised learning.
\newblock {\em Advances in neural information processing systems}, 33:21271--21284, 2020.

\bibitem{simsiam_chen2021exploring}
Xinlei Chen and Kaiming He.
\newblock Exploring simple siamese representation learning.
\newblock In {\em Proceedings of the IEEE/CVF conference on computer vision and pattern recognition}, pages 15750--15758, 2021.

\bibitem{dun2024trustworthy}
Jiale Dun, Jun Wang, Juncheng Li, Qianhui Yang, Wenlong Hang, Xiaofeng Lu, Shihui Ying, and Jun Shi.
\newblock A trustworthy curriculum learning guided multi-target domain adaptation network for autism spectrum disorder classification.
\newblock {\em IEEE Journal of Biomedical and Health Informatics}, 2024.

\bibitem{dermyer2025endodino}
Patrick Dermyer, Angad Kalra, and Matt Schwartz.
\newblock Endodino: A foundation model for gi endoscopy.
\newblock {\em arXiv preprint arXiv:2501.05488}, 2025.

\bibitem{mmpretrain}
Mmpretrain: Openmmlab pre-training toolbox and benchmark.
\newblock 2025.

\bibitem{2_ccauglar2023ulcerative}
{\"U}mit~Mert {\c{C}}a{\u{g}}lar, Alperen {\.I}nci, O{\u{g}}uz Hano{\u{g}}lu, G{\"o}rkem Polat, and Alptekin Temizel.
\newblock Ulcerative colitis mayo endoscopic scoring classification with active learning and generative data augmentation.
\newblock In {\em 2023 IEEE International Conference on Bioinformatics and Biomedicine (BIBM)}, pages 462--467. IEEE, 2023.

\bibitem{singh2021metamed}
Rishav Singh, Vandana Bharti, Vishal Purohit, Abhinav Kumar, Amit~Kumar Singh, and Sanjay~Kumar Singh.
\newblock Metamed: Few-shot medical image classification using gradient-based meta-learning.
\newblock {\em Pattern Recognition}, 120:108111, 2021.

\bibitem{ozdemir2024meta}
Zeynep {\"O}zdemir, Hacer~Yalim Keles, and {\"O}mer~{\"O}zg{\"u}r Tanr{\i}{\"o}ver.
\newblock Meta-transfer derm-diagnosis: Exploring few-shot learning and transfer learning for skin disease classification in long-tail distribution.
\newblock {\em arXiv preprint arXiv:2404.16814}, 2024.

\end{thebibliography}

\end{document}